\documentclass[journal]{IEEEtran}

\ifCLASSINFOpdf
\else
\fi

\usepackage[utf8]{inputenc} % allow utf-8 input
\usepackage[T1]{fontenc}    % use 8-bit T1 fonts
\usepackage{hyperref}       % hyperlinks
\usepackage{url}            % simple URL typesetting
\usepackage{booktabs}       % professional-quality tables
\usepackage{amsfonts}       % blackboard math symbols
\usepackage{nicefrac}       % compact symbols for 1/2, etc.
\usepackage{microtype}      % microtypography
\usepackage{graphicx}
\usepackage{amsmath}
\usepackage{cases}
\usepackage{bm}
\usepackage{color}
\usepackage{multirow}

\begin{document}

\title{Revisiting Multiple Instance Neural Networks}

\author{Xinggang Wang, Yongluan Yan, Peng Tang,~~\IEEEmembership{Student Member, IEEE,}
        Xiang Bai,~\IEEEmembership{Senior Member, IEEE,}
        and Wenyu Liu,~\IEEEmembership{Senior Member, IEEE}% <-this % stops a space
\thanks{X. Wang, Y. Yan, P. Tang, X. Bai, W. Liu were with the School
of Electronics Information and Communications, Huazhong University of Science and Technology, Wuhan, 430074 China e-mail: (\url{xbai@hust.edu.cn}).}% <-this % stops a space
}

\maketitle

\begin{abstract}
  Recently neural networks and multiple instance learning are both attractive topics in Artificial Intelligence related research fields. Deep neural networks have achieved great success in supervised learning problems, and multiple instance learning as a typical weakly-supervised learning method is effective for many applications in computer vision, biometrics, nature language processing, etc.
  In this paper, we revisit the problem of solving multiple instance learning problems using neural networks. Neural networks are appealing for solving multiple instance learning problem.
  The multiple instance neural networks perform multiple instance learning in an end-to-end way, which take a bag with various number of instances as input and directly output bag label. All of the parameters in a multiple instance network are able to be optimized via back-propagation.
  We propose a new multiple instance neural network to learn bag representations, which is different from the existing multiple instance neural networks that focus on estimating instance label.
  In addition, recent tricks developed in deep learning have been studied in multiple instance networks, we find \textit{deep supervision} is effective for boosting bag classification accuracy.
  In the experiments, the proposed multiple instance networks achieve state-of-the-art or competitive performance on several MIL benchmarks. Moreover, it is extremely fast for both testing and training, e.g., it takes only $0.0003$ second to predict a bag and a few seconds to train on a MIL datasets on a moderate CPU.
\end{abstract}


% \fi
\IEEEpeerreviewmaketitle

\section{Introduction}
\label{sec:intro}

Multiple instance learning (MIL) was originally proposed for drug activity prediction \cite{dietterich1997solving}. Now it has been widely applied in many domains and becomes an important problem in machine learning. Many multimedia data have the multiple instance (MI) structure, for example, a text article contains multiple paragraphs, an image can be divided into multiple local regions, and a gene expression data contains multiple genes. MIL is effective to process and understand MI data.

MIL is a kind of weakly-supervised learning (WSL). Each sample is in a form of labeled bags, composed of a wide diversity of instances associated with input features. The aim of MIL, in a binary task, is to train a classifier to predict labels of testing bags, which is based on the assumption that a positive bag contains at least one positive instance while a bag is negative if it is only constituted of negative instances. Thus, the crux of MIL is to deal with the ambiguity of instances labels, especially in positive bags which have plenty of cases with different compositions.

There are many algorithms have been proposed to solve the MIL problem. According to the survey by Amores \cite{amores2013multiple}, MIL algorithms can be divided into three folds: instance-space paradigm, bag-space paradigm and embedded-space paradigm. Instance-space paradigm learns instance classifier and performs bag classification by aggregating the responses of instance-level classifier. Bag-space paradigm exploits bag relations and treats bag as a whole; in particular, bag-to-bag distance/similarity is calculated; then the nearest neighbor or Bayesian classifier is able to do bag classification. Embedded-space paradigm embeds a bag into a vocabulary-based feature space to obtain a compact representation for the bag, e.g., a vector representation; then classical classifiers can be applied to solve the bag classification problem.

Deep neural networks have been applied to solve many machine learning problems. For supervised learning, there are several kinds of neural networks: Deep Belief Networks (DBN) \cite{hinton2006fast} use unsupervised pre-training and take a fixed length vector as input for feature learning and classification; deep Convolutional Neural Networks (CNN) \cite{lecun1998gradient, Ref:Krizhevsky2012} take 2D image as input and have dominated image recognition; deep Recurrent Neural Networks (RNN) \cite{williams1989learning} and Long Short Term Memory (LSTM) networks \cite{hochreiter1997long} take sequential data as input, such as text and speech, and are good at dealing with sequential prediction.
Usually, training these deep networks requires a large number of fully labeled data, i.e., each instance requires a label.
However, in MIL, only bag labels can be got. Meanwhile, MI data have a more complex structure which is a set of instances. The numbers of instances are different for different bags. These problems make it hard to deal with MIL problem by conventional neural networks.

Before the raising of deep learning, there were some research works trying to solve the MIL problem using neural networks. Ramon and Raedt \cite{ramon2000multi} firstly proposed a multiple instance neural network (MINN). The network estimates instance probabilities before the last layer and calculates bag probability using a convex max operator (i.e., log-sum-exp). The network can be trained using back-propagation. Zhang and Zhou \cite{zhou2002neural} also proposed a multiple instance network which calculates bag probability by directly taking the max of instance probabilities.

A MINN takes a various number of instances as input. For each instance, its representation is gradually learned layer by layer guided by multiple instance supervision.
To inject multiple instance supervision, there are two different network architectures. Following the naming style in a classical MIL work \cite{andrews2002support}, we name the two networks as mi-Net and MI-Net, which aim at the instance-space paradigm and embedded-space paradigm \cite{amores2013multiple} respectively. In mi-Net, there are instance classifiers in the each layer. We are able to obtain instance labels for both training and testing bags, which is an appealing property in some applications. While in MI-Net, there is no instance classifier. It directly builds a fixed-length vector as the bag representation and then learns bag classifier. Compared with mi-Net, MI-Net can obtain better bag classification accuracy. The previous works are in the category of mi-Net. We newly propose MI-Net in this paper.

A key component in MINN is MIL Pooling Layer (MPL), which aggregates either instance probability distribution vectors or instance feature vectors into a bag feature vector. It bridges MI data with conventional neural networks. Since it must be differentiable, there are a few choices, such as max pooling, mean pooling, and log-sum-exp pooling. These pooling methods are compared and discussed in the experiments part. Besides of MIL pooling layer, we use fully connected layers with non-linear activations for instance feature learning. In MIL benchmarks, instance features are hand-crafted and raw data of instances are given. Even so, it is beneficial to do feature transformation guided by the supervision of bag labels. In the last of MI-Net, we use a fully connected layer with only one neuron to match the predicted bag label with ground-truth in training.

Training neural networks using complex MI data is a challenging task. To learn good instance feature, we have tried to adopt various recent progress of deep learning in MINN, such as dropout \cite{srivastava2014dropout}, ReLU \cite{nair2010rectified}, deeply supervised nets (DSN) \cite{lee2015deeply} and Residual Connections \cite{he2015deep}. We find DSN is the most effective one. This is due to DSN is able to better use hierarchical features in networks. Also, residual connections do a great job in networks.

To summarize, we revisit the problem of solving multiple instance learning using neural networks. This branch of MIL algorithm is ignored by current MIL research community. But it is highly effective and efficient. Different from most MIL algorithms, it is able to learn instance features in an end-to-end manner.
This paper focuses on neural networks for end-to-end MIL with comprehensive studies on MIL benchmarks. The main contributions of this paper include two extremely fast and scalable methods for MIL, i.e., mi-Net and MI-Net, and introducing deep supervision and residual connections for MIL.

We organize the rest of this paper as follow. Section~\ref{sec:related_work} briefly reviews previous works on MIL. In Section ~\ref{sec:form_mil_net}, we propose end-to-end MIL networks. Our experimental results are presented on several MIL benchmarks in Section~\ref{sec:exp}. Some discussions of experimental setups are presented in Section~\ref{sec:discuss}. Finally, in Section~\ref{sec:conclu} we conclude the paper with some future works.

\section{Related Work}
\label{sec:related_work}

Previous works on solving MIL using neural networks include \cite{ramon2000multi, zhou2002neural, zhang2004improve, zhang2004ensembles}. \cite{ramon2000multi} introduced to use a log-sum-exp as the convex max to calculate bag probabilities from instance probabilities. \cite{zhou2002neural} changed to a different loss function and directly applied max function. \cite{zhang2004improve} improved multiple instance neural networks by feature selection using Diverse Density and PCA. \cite{zhang2004ensembles} showed that ensemble methods could be integrated with multiple instance neural networks. Then, solving MIL using neural networks has been ignored in machine learning research. This paper revisits this problem, proposes new network structures, and investigates recent neural network tricks.

Multiple Instance Learning (MIL) has received a lot of attentions since it helps to solve a range of real applications. Till now, lots of MIL methods have been proposed to either develop effective MIL solvers or apply MIL to solve application problems. A comprehensive survey of MIL algorithms and applications can be found in \cite{amores2013multiple}. Here, we focus on give a brief review of the most recent MIL algorithm, especially the ones related to deep neural networks and feature learning.

From the view of embedded-space paradigm for MIL, the most recent method is the scalable MIL algorithm, i.e., solving MIL using Fisher Vector (FV) coding \cite{Sanchez2013}, which is called miFV \cite{Ref:Wu2015}. miFV transforms instance feature into high-dimensional space using a pre-trained Gaussian mixture model and FV coding. The proposed MI-Net learns instance feature using deep multiple instance supervision. And MI-Net achieves better bag classification accuracy and is much faster than miFV.

The idea of using neural networks for solving MIL problem has been studied in some computer vision studies, such as \cite{Ref:Wu2015, pinheiro2015image}. Wu et. al \cite{Ref:Wu2015} proposed deep MIL which uses max pooling to find positive instances/patches for image classification and annotation. Pinheheiro et. al \cite{pinheiro2015image} used log-sum-exp pooling in deep CNN for weakly supervised semantic segmentation. The proposed mi-Net follows the path of these two works; different from them, mi-Net utilizes deep supervision, and focuses on more general MIL problems. Besides of integrating MIL into deep neural networks, Wang et. al proposed a method to combine MIL with support vector machine using a relaxed MIL constraint \cite{wang2015relaxed} and applied this for object discovery. However, they pay more attention on vision applications (e.g., image classification, image annotation, and semantic segmentaion, etc.), which are based on convolutional image features. Meanwhile, they always finetune neural network models pre-trained on other much larger datasets like ImageNet~\cite{Ref:Deng2009}. Moreover, they also only focus more on instance-space MIL.Compared with theirs, we focus on appling MIN structure for more general MIL problems. Notice that for general MIL problems, there are no available large datasets for pre-training like computer vision, which makes it more difficult to train MINN efficiently. We have shown many tricks to train our networks from scratch on MIL benchmarks with limited training data, and achieved many inspiring results. Meanwhile, we have investigated both mi-Net and MI-Net, and experiments have shown that the MI-Net outperforms mi-Net in more cases.

\section{Multiple Instance Neural Networks}
\label{sec:form_mil_net}

In this section, we will firstly introduce the formulation of MIL, then give various networks for MIL, and lastly study the MIL pooling methods and training loss.

\subsection{Notations}

Here we first review the definition of MIL.
Given a set of bags ${X} = \{X_{1}, X_{2}, ..., X_{N}\}$ and instance features of $i th$ bag $X_{i} = \{{x}_{i1}, {x}_{i2}, ..., {x}_{im_{i}}\}, {x}_{ij} \in \mathbb{R}^{d \times 1}$,
where $N$ and $m_{i}$ denote the number of bags and the number of instances in bag $X_{i}$ respectively.
Suppose $Y_{i} \in \{0, 1\}$ and $y_{ij} \in \{0, 1\}$ are the label of bag $X_{i}$ and instance $x_{ij}$ separately, where $1$ means positive and $0$ means negative.
In MIL, only bag labels are given during training, and there are two MIL constraints:
\begin{itemize}
\item If bag $X_{i}$ is negative, then all instances in $X_{i}$ will be negative,
i.e., if $Y_{i} = 0$, then all $y_{ij} = 0$;
\item If bag $X_{i}$ is positive, then at least one instance in $X_{i}$ will be positive,
i.e., if $Y_{i} = 1$, then $\mathop \sum \limits_{j=1}^{m_{i}} y_{ij} \geq 1$.
\end{itemize}

Since instance label is not given in training phase, solving the MIL problem is challenging. In MINNs, there are two strategies: the first one is to infer instance label in the network, i.e., placing instance probabilities of being positive as a hidden layer in the network; the second one is to use learn bag representation in the network and directly carry out bag classification without calculating instance probability. The first strategy has been studied in \cite{ramon2000multi, zhou2002neural, Ref:Wu2015}. The second strategy is newly proposed in this paper. In the following sub-sections, we will give the descriptions of MINNs.

Let us consider a setting of a single bag $X_i$ with multiple instances $x_{ij}$ that is passed through a MINN. A MINN is made out of $L$ layers, each of which consists of a non-linear transformation $H^\ell(\cdot)$, where $\ell$ indexes the layer. $H^\ell(\cdot)$ can be a composite of operations such as inner product (or fully connection), or rectified linear units (ReLU) \cite{glorot2011deep}. We denote the output of the $\ell^{th}$ layer of an instance $x_{ij}$ as $x^\ell_{ij}$.

\subsection{mi-Net: Instance-Space MIL Algorithm}
\label{sec:instance}

\begin{figure*}[t]
  \centerline{
    \includegraphics[width=15cm]{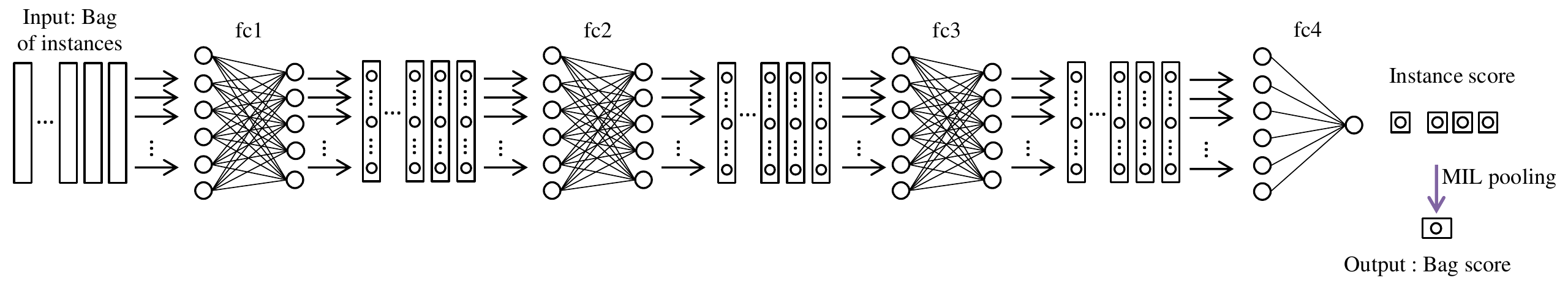}
  }
  \caption{A mi-Net with four fully connected layers. The number of output of fully connected layers are 256, 128, 64 and 1 respectively. The last layer is a MIL pooling layer with instance probabilites as input and bag probability as output.}
  \label{fig:net1}
\end{figure*}

At first, we review traditional multiple instance neural networks \cite{ramon2000multi, zhou2002neural, Ref:Wu2015}, which are named as mi-Net. As shown in Fig.~\ref{fig:net1}, each instance in a bag is first fed into several fully connected (fc) layers with activation function (in this paper we use four fc layers and ReLU activation). Thus, we get the instance feature denoted as $x^{L-2}_{ij}$ in the $(L-2)th$ layer and instance probability denoted as $p^{L-1}_{ij}$. $p^{L-1}_{ij}$ is a scalar in the  range of $[0,1]$. In the last layer, there is a MIL pooling layer (described in Section~\ref{sec:pooling_methods}) which takes instance probabilities as input and outputs bag probability, denoted as $P^{L}(X_i)$.

These first $L-1$ fc layers can learn some more semantic instance features compared with original ${x}_i{ij}$ (higher layer corresponding to higher semantic features).
After learning these instance features, a fc layer which only has one neuron with sigmoid activation, is used to predict the positiveness of instances.

But unlike traditional neural networks, for mi-Net, we only have bag labels for training but instance labels are not available. To address this problem, we treat the instance labels as latent variables and infer them during the network training. We design a layer to aggregate instance scores into bag score. Here, a MIL pooling layer is used to aggregate these instance scores into the final the positiveness of bag.

The MIL pooling method satisfies the MIL constraints: If a bag is positive, there should have at least one instance with large positiveness. Otherwise, all instances in the bag should have low positiveness. Since the pooling layer is integrated into the neural network, the pooling function should be differentiable. There typical MIL pooling will be introduced in Section~\ref{sec:pooling_methods}.

In summary, the mi-Net can be formulated as:

\begin{equation}
    \label{equ:mi-Net}
    \begin{cases}
         x^\ell_{ij} = H^\ell (x^{\ell-1}_{ij}), \\
         P^{L}_i = M^L(p^{L-1}_{ij|j=1\dots m_i}).
    \end{cases}
\end{equation}

\subsection{MI-Net: A new Embedded-Space MIL Algorithm}

We propose a series of new multiple instance neural networks which do not rely on inferring instance probability. The networks directly learn bag representation and produce better bag classification accuracy. These methods belong to the category of embedded-space MIL algorithms defined in the survey \cite{amores2013multiple}. Following the naming style in \cite{andrews2002support}, we name this networks as MI-Net.

\begin{figure*}[t]
  \centerline{
    \includegraphics[width=15cm]{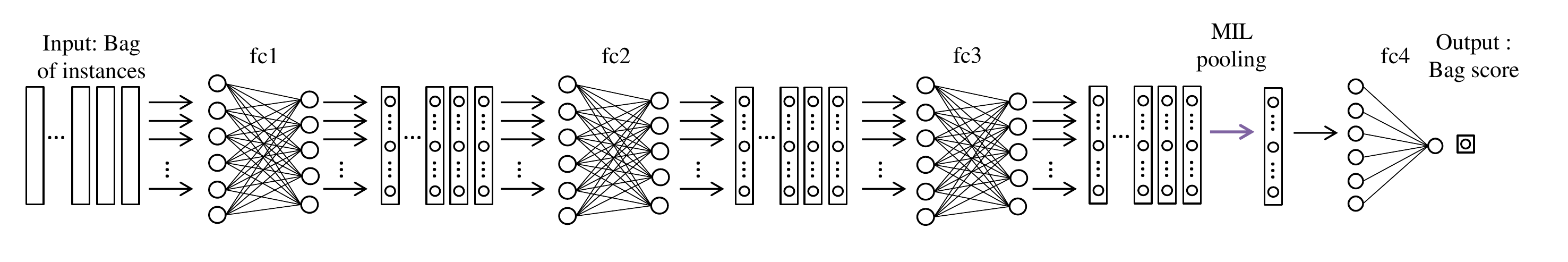}
  }
  \caption{The proposed MI-Net with three fully connected layers and one MIL pooling layer. The number of output of fully connected layers are 256, 128 and 64 respectively.}
  \label{fig:net2}
\end{figure*}

In Figure~\ref{fig:net2}, we show a plain MI-Net with three fully connected layer and one MIL pooling layer. The change of network structure leads the network to focus on learning bag representation, rather than predicting instance probability. No matter how many input instances there are, the MIL pooling layer aggregates them into one feature vector as a bag representation. 
At last, a fc layer with only one neuron and sigmoid activation takes the bag representation as input and predicts bag probability. This plain MI-Net is formulated as:

\begin{equation}
    \label{equ:MI-Net}
    \begin{cases}
         x^\ell_{ij} = H^\ell (x^{\ell-1}_{ij}), \\
         X^{\ell}_i = M^\ell(x^{\ell-1}_{ij|j=1\dots m_i}).
    \end{cases}
\end{equation}

\subsection{MI-Net with Deep Supervision}

\begin{figure*}[t]
  \centerline{
    \includegraphics[width=15cm]{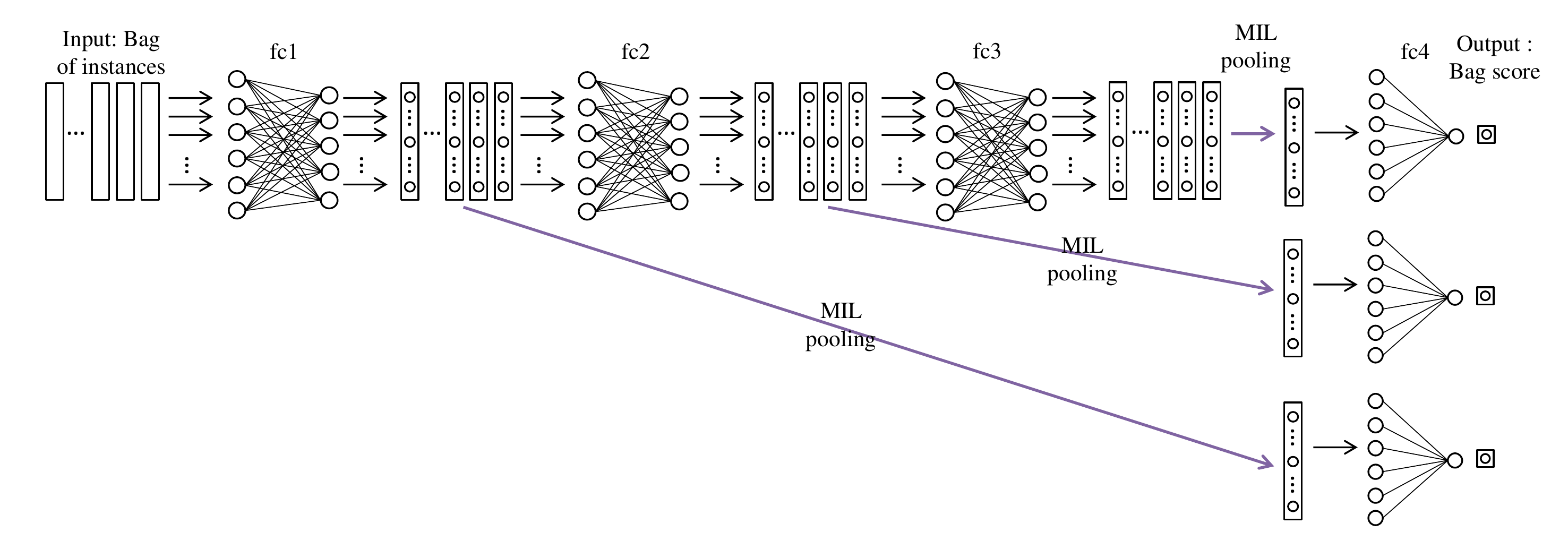}
  }
  \caption{The proposed MI-Net with deep supervision. There are three fully connected layers for learning instance features which are in the size of 256, 128 and 64 respectively. And there are three MIL pooling layers for generating bag feature and the bag features are connected to the bag label via a fully connected layer with one neuron respectively.}
  \label{fig:net3}
\end{figure*}

Inspired by the Deeply-Supervised Nets (DSN)~\cite{lee2015deeply}, we add deep supervisions in MI-Net as shown in Figure~\ref{fig:net3}.
That is, for each middle fc layer that can learn instance features, a fc layer for predicting instance scores with a MIL pooling layer follows it.
During training, the supervision is added to each level.
And during testing, we compute the mean score for each level.
The MI-Net with deep supervision is formulated as:
\begin{equation}
    \label{equ:ds}
    \begin{cases}
         x^\ell_{ij} = H^\ell (x^{\ell-1}_{ij}), \\
         X^{\ell, k}_i = M^\ell(x^{k}_{ij|j=1\dots m_i}), k \in \{1,2,3\},
    \end{cases}
\end{equation}
where the index $k$ in $X^{\ell, k}_i$ means we learn multiple bag features from all different levels of instance features by MIL pooling. 
MI-Net with deep supervision is able to utilize  multiple hierarchies to get better bag classification accuracy. It can be interpreted from two folds: (1) In training instance feature in bottom layers can receive better supervision; and (2) in testing, we can average multiple bag probabilities  to get a more robust bag label. In this paper, we set the weights of different levels equally.

\subsection{MI-Net with Residual Connections}
  \label{ssec:residual}
\begin{figure*}[t]
  \centerline{
    \includegraphics[width=15cm]{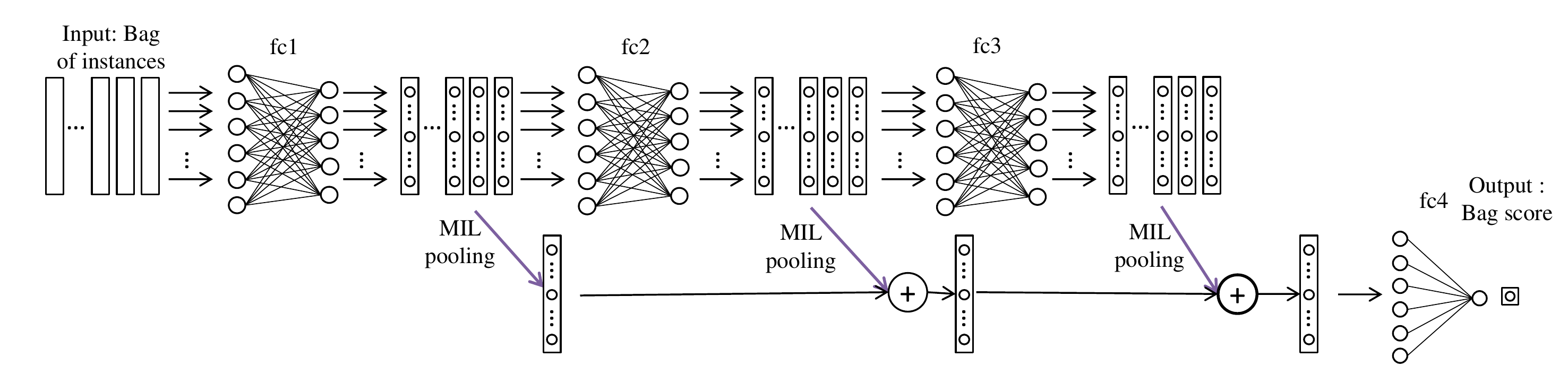}
  }
  \caption{The proposed MI-Net with residual connections. The first fully connected layer produces a bag feature vector. The latter fully connected layers learn the residuals of bag representation. The size of fully connected layers are all 128.}
  \label{fig:net4}
\end{figure*}

Recently, deep residual learning was proposed in \cite{he2015deep} and showed the impressive improvement in image recognition by utilizing very deep neural networks.
We study the residual connections in MI-Net as shown in Figure~\ref{fig:net4}.
MI-Net with residual connections are formulated as:
\begin{equation}
    \label{equ:ds}
    \begin{cases}
         x^\ell_{ij} = H^\ell (x^{\ell-1}_{ij}), \\
         X^{1}_i = M^\ell(x^{1}_{ij|j=1\dots m_i}), \\
         X^{\ell}_i = M^\ell(x^{\ell}_{ij|j=1\dots m_i}) + X^{\ell-1}, \ell > 1.
    \end{cases}
\end{equation}
Different from the original residual learning in \cite{he2015deep} which learns representation residuals using convolution, batch normalization and ReLU, we learn the bag representation residuals via fully connected layers, ReLU and MIL pooling. In the end of the network, final bag representation is connected to the bag label via a fc layer with one neuron and sigmoid activation.

% \begin{equation}
%     \label{equ:prob_instance}
%     P_{i1} = \frac{1}{1 + \exp(-S_{i1})}, \ \ P_{i0} = 1 - P_{i1}.
% \end{equation}

\subsection{MIL Pooling Methods}
\label{sec:pooling_methods}

As referred before, we use a MIL pooling layer to get patch scores or patch representations.
In this paper, we use three popular used MIL pooling methods: max pooling, mean pooling, and log-sum-exp (LSE) pooling, as shown in Eq.~(\ref{equ:pooling}), where $f_{i}$ is the input, $o$ is the output, $m$ is the number of input, and $r$ is a hyper-parameter.
All these methods satisfy the constraints referred in Section~\ref{sec:instance}.
Actually the LSE~\cite{boyd2004convex} is a smooth version and convex approximation of the max function.
The hyperparameter $r$ controls how the smoothness of approximation.
That is, it is more approximate to max when $r$ is large and more approximate to mean when $r$ is small.

\begin{equation}
    \label{equ:pooling}
    \begin{cases}
         \mathrm{max}:\, & M^\ell(x^{\ell-1}_{ij|j=1\dots m_i}) =  \mathop{\max} \limits_{j} x^{\ell-1}_{ij}, \\
        %\label{equ:mean_pooling}
         \mathrm{mean}:\, & M^\ell(x^{\ell-1}_{ij|j=1\dots m_i}) =  \frac{1}{m_i} \mathop{\sum} \limits_{j=1}^{m_i} x^{\ell-1}_{ij}, \\
        %\label{equ:LSE}
         \mathrm{LSE}:\, & M^\ell(x^{\ell-1}_{ij|j=1\dots m_i}) =  r^{-1} \log [\frac{1}{m_i} \mathop{\sum} \limits_{j=1}^{m_i} \exp(r \cdot x^{\ell-1}_{ij})].
    \end{cases}
\end{equation}

% \begin{equation}
%     \label{equ:mean_pooling}
%     o = \frac{1}{n} \mathop{\sum} \limits_{i=1}^{m} f_{i}.
% \end{equation}

% \begin{equation}
%     \label{equ:LSE}
%     o = r^{-1} \log [\frac{1}{n} \mathop{\sum} \limits_{i=1}^{m} \exp(r \cdot f_{i})].
% \end{equation}

\subsection{Training Loss}

For both mi-Net and MI-Net, we can get the bag scores.
Here we will define the loss function during training.
As we are aiming at predicting labels of bags, it is natural to choose the cross entropy loss function, as in Eq.~(\ref{equ:loss}), where $S_i$ is the bag score of $i$ bag.
This loss is added to each bag scores level for deep supervision.

\begin{equation}
    \label{equ:loss}
    % \mathrm{Loss}(\mathbf{S}_{i}, Y_{i}) = - \{(1 - Y_{i}) \log p_{i0} + Y_{i} \log p_{i1}\}.
	\mathrm{Loss}(S_{i}, Y_{i}) = - \{(1 - Y_{i}) \log(1 - S_{i}) + Y_{i} \log S_{i}\}.
\end{equation}

As all parts of our multiple instance network are differentiable, we can train these networks by standard back-propagation with Stochastic Gradient Descent (SGD).

\begin{table*}  
  \caption{Detailed characteristics of the datasets. "\# positive" ("\#negative")  presents the number of positive(negative) bags used in each round. For Text category dataset, because it contains 20 sub-datasets, we present three of them in it.}
  \label{table:exp0}
  \centering 
  \begin{tabular}{lcccccccc}
    \toprule
    \multirow{2}{*}{\# dataset} & \multirow{2}{*}{\# attribute} & \multicolumn{3}{c}{\# bag} & \multicolumn{3}{c}{\# instance} \\
    \cmidrule(lr){3-5}
    \cmidrule(lr){6-8}
     &  & positive & negative & total & min & max & total  \\
    \midrule
 	MUSK1 & $166$ & $47$ & $45$ & $92$ & $2$ & $40$ & $476$\\
	MUSK2 & $166$ & $39$ & $63$ & $102$ & $1$ & $1044$ & $6598$\\
	Elephant & $230$ & $100$ & $100$ & $200$ & $3$ & $13$ & $1391$ \\
	Fox & $230$ & $100$ & $100$ & $200$ & $2$ & $13$ & $1320$ \\
	Tiger & $230$ & $100$ & $100$ & $200$ & $1$ & $13$ & $1220$ \\
	Text(Zhou) alt.atheism & $200$ & $50$ & $50$ & $100$ & $22$ & $76$ & $5443$ \\
	Text(Zhou) comp.graphics & $200$ & $49$ & $51$ & $100$ & $12$ & $58$ & $3094$ \\
	Text(Zhou) comp.os.ms-windows.misc & $200$ & $50$ & $50$ & $100$ & $25$ & $82$ & $5175$ \\
    \bottomrule             
\end{tabular}
\end{table*}

\begin{table*}[t]
  \caption{Average prediction accuracy (in $\%$) of different methods for bag classification on five MIL benchmarks.}
  \label{table:exp1}
  \centering
  \begin{tabular}{llllll}
    \toprule
    Name     & MUSK1 & MUSK2 & Elephant  & Fox   & Tiger \\
    \midrule
     mi-SVM~\cite{andrews2002support}     & $0.780$      & $0.702$ & $0.822$ & $0.582$ & $0.784$\\
     MI-SVM~\cite{andrews2002support}     & $0.779$      & $0.843$ & $0.814$ & $0.578$ & $0.840$ \\
     EM-DD~\cite{zhang2001dd} & $0.849$ & $0.869$ & $0.771$ & $0.609$ & $0.730$ \\
     MI-Kernel~\cite{gartner2002multi} & $0.880$ & $0.893$ & $0.843$ & $0.603$ & $0.842$\\
     MI-Graph~\cite{zhou2009multi} & $0.900$ & $0.900$ & $0.851$ & $0.612$ & $0.819$ \\
     mi-Graph~\cite{zhou2009multi} & $0.889$ & $\mathbf{0.903}$ & $0.868$ & $0.616$ & $\mathbf{0.860}$ \\
     miVLAD~\cite{wei2016scalable}     & $0.871$      & $0.872$ & $0.850$ & $0.620$ & $0.811$ \\
     miFV~\cite{wei2016scalable}       & $\mathbf{0.909}$ & $0.884$ & $0.852$ & $0.621$ & $0.813$ \\
    \midrule
     mi-Net    			& $0.889$ & $0.858$ & $0.858$ & $0.613$ & $0.824$ \\
     MI-Net    			& $0.887$ & $0.859$ & $0.862$ & $0.622$ & $0.830$ \\
	MI-Net with DS		& $0.894$ & $0.874$ & $\mathbf{0.872}$ & $\mathbf{0.630}$ & $0.845$ \\
	MI-Net with RC		& $0.898$ & $0.873$ & $0.857$ & $0.619$ & $0.836$ \\
    \bottomrule
  \end{tabular}
\end{table*}

\section{Experiments}
\label{sec:exp}

In this section, we perform experiments to test mi-Net, MI-Net and its variations on different MIL benchmarks, including molecule activity, image, and text categorization.

\subsection{Datasets}

We test these methods on three widely-used MILbenchmarks in different applications, including drug activation prediction, automatic image annotation and text categorization.
For evaluation, we run five times 10-fold cross validation and report the average results.

\paragraph{Drug Activation Prediction}

MUSK~\cite{dietterich1997solving} datasets are used to predict whether a drug molecule can bind well to target protein.
Each molecule is exhibited as multiple shapes, which are described as $166$-dimension features.
In the MIL problem, we can regard a molecule as a bag and represent different shapes belonging to the same molecule as instances of this bag.
$476$ instances are included in MUSK1 which is divided into $47$ positive bags and $45$ negative bags, while $6598$ instances are included in MUSK2 which is divided into $39$ positive bags and $63$ negative bags.

\paragraph{Automatic Image Annotation}

The Elephant, Fox and Tiger datasets~\cite{andrews2002support}, are all composed of $100$ positive bags from the target class animal images and $100$ negative bags randomly chosen from other class animal images.
Here, an image is represented as a bag, which contains a set of regions we called instances in MIL problems.
When searching for a target object, we use this network to obtain the keywords of images.
Moreover, each image is represented by $2$ to $13$ instances which are $230$-dimension features that describe the color, texture, and shape in regions of an image.

\paragraph{Text Categorization}

Besides the above datasets, the text categorization is another widely used application of MIL problems.
Here, we take twenty datasets derived from the 20 Newsgroups corpus~\cite{zhou2009multi}.
In each category, $100$ bags are included among which half bags are positive and the rest of bags are negative.
Each positive bag contains $3$\% posts from the target class and the rest from other categories, while the instances of negative bags are all randomly drawn from other categories.
In addition, each instance is represented by the top $200$ TF-IDF features.

Detailed characteristics of these datasets are summarized in Table~\ref{table:exp0}.

\subsection{Experimental Setup}
These neural networks contain four fully connected (fc) layers and first three fc layers are followed by a dropout layer ($0.5$ dropout ratio).
As referred in Section~\ref{sec:form_mil_net}, we present the performance of the proposed multiple instance learning approaches: (1) mi-Net:We learn instance scores from four fc layers and aggregate instance scores into bag scores to predict the label of the bag via MIL pooling layer.
(2) MI-Net: Input instances are aggregated into bag representation by first three fc layers and MIL pooling layer, and then use the last fc layer to predict bag probability.
(3) MI-Net with Deep Supervision (MI-Net with DS): Different from MI-Net, each middle fc layer is followed by a MIL pooling layer and fc layer to compute bag scores. 
The loss function of MI-Net with DS sums up all middle entropy losses to do backpropagation with SGD for training, and the average of each bag score is used for testing.
(4) MI-Net with Residual Connections (MI-Net with RC): Residual connections are built between each middle bag representation, and followed by a fc layer to obtain bag score.

As for the numbers of neurons in fc layers, there are $256$, $128$, $64$, $1$ in mi-Net, MI-Net and MI-Net with DS while $128$, $128$, $128$, $1$ in MI-Net with RC.
Weights of fc layers are all initialized using a glorot-uniform distribution~\cite{glorot2010understanding}.
Biases are all initialized to be 0.
For different datasets, the learning rate, weight decay and momentum are set suitable values that you can find in the configuration file of our code.
All networks are trained with SGD, and one bag is inputted as a batch for training and testing.
Moreover, about training and testing time, e.g., it takes only $0.0003$ second to predict a bag and $0.0008$ second to train on MUSK1 dataset on a moderate CPU.
Our code is written in Python, based on Keras~\cite{chollet2015keras}, and all of our experiments are running on a PC with Inter(R) i7-4790K CPU (4.00GHZ) and 32GB RAM.
The code for reproducing results will be available upon acceptance.

\begin{table*}[t]
  \caption{Average prediction accuracy (in $\%$) for bag classification on text categorization.}
  \label{table:exp2}
  \centering
  \begin{tabular}{llllllll}
    \toprule
    Dataset                 & MI-Kernel \cite{gartner2002multi}  & miGraph \cite{zhou2009multi}  & miFV \cite{wei2016scalable}    & mi-Net  & MI-Net & MI-Net with DS & MI-Net with RC\\
    \midrule
     alt.atheism            & $0.602$     & $0.655$   & $0.848$   & $0.758$  & $0.776$  & $\mathbf{0.860}$  & $0.858$\\%164
     comp.graphics          & $0.470$     & $0.778$   & $0.594$   & $\mathbf{0.830}$  & $0.826$  & $0.822$  & $0.828$\\%165
     comp.windows.misc      & $0.510$     & $0.631$   & $0.615$   & $0.658$  & $0.678$  & $0.716$  & $\mathbf{0.720}$\\%166 
     comp.ibm.pc.hardware   & $0.469$     & $0.595$   & $0.665$   & $0.772$  & $0.778$  & $\mathbf{0.792}$  & $0.784$\\%167
     comp.sys.mac.hardware  & $0.445$     & $0.617$   & $0.660$   & $0.746$  & $0.792$  & $0.794$  & $\mathbf{0.810}$\\%168
     comp.window.x          & $0.508$     & $0.698$   & $0.768$   & $0.746$  & $0.786$  & $0.812$  & $\mathbf{0.820}$\\%169
     misc.forsale           & $0.518$     & $0.552$   & $0.565$   & $0.580$  & $0.652$  & $0.686$  & $\mathbf{0.696}$\\%170
     rec.autos              & $0.529$     & $0.720$   & $0.667$   & $0.746$  & $0.774$  & $0.776$  & $\mathbf{0.792}$\\%171
     rec.motorcycles        & $0.506$     & $0.640$   & $0.802$   & $0.716$  & $0.762$  & $\mathbf{0.868}$  & $0.856$\\%172
     rec.sport.baseball     & $0.517$     & $0.647$   & $0.779$   & $0.808$  & $0.856$  & $0.874$  & $\mathbf{0.880}$\\%173
     rec.sport.hockey       & $0.513$     & $0.850$   & $0.823$   & $0.860$  & $0.862$  & $0.912$  & $\mathbf{0.918}$\\%174
     sci.crypt              & $0.563$     & $0.696$   & $0.760$   & $0.608$  & $0.694$  & $\mathbf{0.812}$  & $0.796$\\%175
     sci.electronics        & $0.506$     & $0.871$   & $0.555$   & $0.932$  & $0.930$  & $0.926$  & $\mathbf{0.938}$\\%176
     sci.med                & $0.506$     & $0.621$   & $0.783$   & $0.792$  & $0.818$  & $\mathbf{0.848}$  & $0.842$\\%177
     sci.space              & $0.547$     & $0.757$   & $\mathbf{0.818}$   & $0.694$  & $0.752$  & $\mathbf{0.818}$  & $0.810$\\%178
     soc.religion.christian & $0.492$     & $0.590$   & $0.814$   & $0.718$  & $0.782$  & $0.820$  & $\mathbf{0.822}$\\%179
     talk.politics.guns     & $0.477$     & $0.585$   & $0.747$   & $0.596$  & $0.652$  & $\mathbf{0.780}$  & $0.762$\\%180
     talk.politics.mideast  & $0.559$     & $0.736$   & $0.793$   & $0.774$  & $0.794$  & $\mathbf{0.842}$  & $0.824$\\%181
     talk.politics.misc     & $0.515$     & $0.704$   & $0.697$   & $0.602$  & $0.654$  & $\mathbf{0.776}$  & $0.736$\\%182
     talk.religion.misc     & $0.554$     & $0.633$   & $0.739$   & $0.700$  & $0.700$  & $0.758$  & $\mathbf{0.764}$\\%183
    \midrule
    average                 & $0.515$     & $0.679$   & $0.726$   & $0.737$  & $0.766$  & $\mathbf{0.815}$  & $0.813$\\
    \bottomrule
  \end{tabular}
\end{table*}

\subsection{Experimental Results}

Experimental results are shown in Table~\ref{table:exp1} and Table~\ref{table:exp2}.
The best performance of each dataset is bolded.
Notice that using different pooling methods for these networks will produce different results for each dataset.
Here, we choose the best one as the final result (for text categorization, the max pooling achieves the best performance consistently).
And we will discuss the influence of pooling methods later.
Particularly, it achieves state-of-the-art performance on Elephant, Fox, and text categorization, and nearly best accuracies on other datasets.
These results demonstrate the effectiveness of these multiple instance networks.
From these results, we can observe that these networks achieve highly competitive results.

We can easily find that the embedded-space network MI-Net seems more competitive than the instance-space network mi-Net, which is consistent with other MIL algorithms.
In five benchmark datasets, MI-Net with DS achieves almost all best results than other methods, which verifies the network with deep supervision will be more robust to predict bag label.
Additionally, MI-Net with RC also gets good results on these five benchmark datasets.
In text categorization datasets, MI-Net with DS achieves the superior performance and results of MI-Net with RC is slightly worser than results of MI-Net.
The average accuracy of all 20 datasets as evaluation indicates that MI-Net and its two variations outperform other five competing algorithms, including MI-Kernel \cite{gartner2002multi}, miGraph \cite{zhou2009multi}, miFV \cite{wei2016scalable} and mi-Nets.

\section{Discussion}
\label{sec:discuss}
In this section, we discuss the influence of different pooling methods, deep supervision, residual connections on the networks.
The width and depth of networks which may have impact on the performance is also considered in discussion.

\begin{table*}[ht]
  \caption{The influence of different pooling methods for MI-Net with DS on five MIL benchmarks.}
  \label{table:bag_pooling}
  \centering
  \begin{tabular}{llllll}
    \toprule
    Pooling Method   & MUSK1  & MUSK2  & Elephant  & Fox  & Tiger \\
    \midrule
    max    & $\mathbf{0.894}$ & $\mathbf{0.874}$ & $\mathbf{0.870}$ & $\mathbf{0.630}$ & $0.826$ \\
    mean    & $0.886$ & $0.858$ & $0.867$ & $0.615$ & $\mathbf{0.845}$ \\
    LSE    & $0.891$ & $0.874$ & $0.872$ & $0.625$ & $0.840$ \\
    \bottomrule
  \end{tabular}
\end{table*}

\begin{table*}[ht]
  \caption{The influence of deep supervision for MI-Net on five MIL benchmarks, where DS means deep supervision.}
  \label{table:dsn}
  \centering
  \begin{tabular}{llllll}
  \toprule
  Method    & MUSK1 & MUSK2 & Elephant  & Fox   & Tiger \\
  \midrule
  MI-Net with DS      & $\mathbf{0.894}$  & $\mathbf{0.874}$ & $\mathbf{0.870}$ & $\mathbf{0.630}$ & $\mathbf{0.845}$ \\
  MI-Net without DS   & $0.887$  & $0.859$ & $0.862$ & $0.622$ & $0.830$ \\
  \bottomrule
 \end{tabular}
\end{table*}

\begin{table*}[ht]
  \caption{The influence of residual connections for MI-Net on five MIL benchmarks, where RC means residual connections.}
  \label{table:rc}
  \centering
  \begin{tabular}{llllll}
    \toprule
     Method	& MUSK1 & MUSK2 & Elephant & Fox & Tiger \\
    \midrule
     MI-Net with RC		& $\mathbf{0.898}$  & $\mathbf{0.873}$ & $0.857$ & $0.619$ & $\mathbf{0.836}$ \\
     MI-Net without RC	& $0.887$  & $0.859$ & $\mathbf{0.862}$ & $\mathbf{0.622}$ & $0.830$ \\
    \bottomrule
  \end{tabular}
\end{table*}

\subsection{The Influence of Different Pooling Methods}

There are three pooling methods applied to these networks, including max pooling, mean pooling and LSE pooling.
As referred in Section~\ref{sec:form_mil_net}, in embedded-space, instance features of the same bag are aggregated into the bag representation through pooling methods;
in instance-space, instance scores of the same bag are aggregated into bag scores.
We test the influence of different pooling methods on MI-Net with DS.
From Table~\ref{table:bag_pooling}, we can observe that max pooling is preferable compared with other methods.

\subsection{The Influence of Deep Supervision}

To illustrate the effectiveness of deep supervision, we compare our MI-Net with deep supervision to the network without deep supervision, which only do MIL pooling and bag score prediction on the third fc layer.
The effectiveness of deep supervision is validated on five MIL benchmark datasets, as shown in Table~\ref{table:dsn}.
From the results, we can observe that the performance is boosted by deep supervision for all datasets and networks.
Deep supervision is essential for learning good instance features in multiple instance networks.

\subsection{The Influence of Residual Connections}

In order to show the improvement of residual connections, MI-Net with Residual Connections which learns the bag representation residuals, is compared to MI-Net.
As referred in Section~\ref{table:rc}, the influence of residual connections is proved on five MIL benchmark datasets.
The results of MI-Net with Residual Connections are better than MI-Net without Residual Connections, except for Elephant and Tiger.
Residual Connections may also have a positive impact on learning good bag representation in multiple instance networks.

\subsection{The Influence of network depth and width}
\begin{figure}[t]
  \centerline{
    \includegraphics[width=8cm]{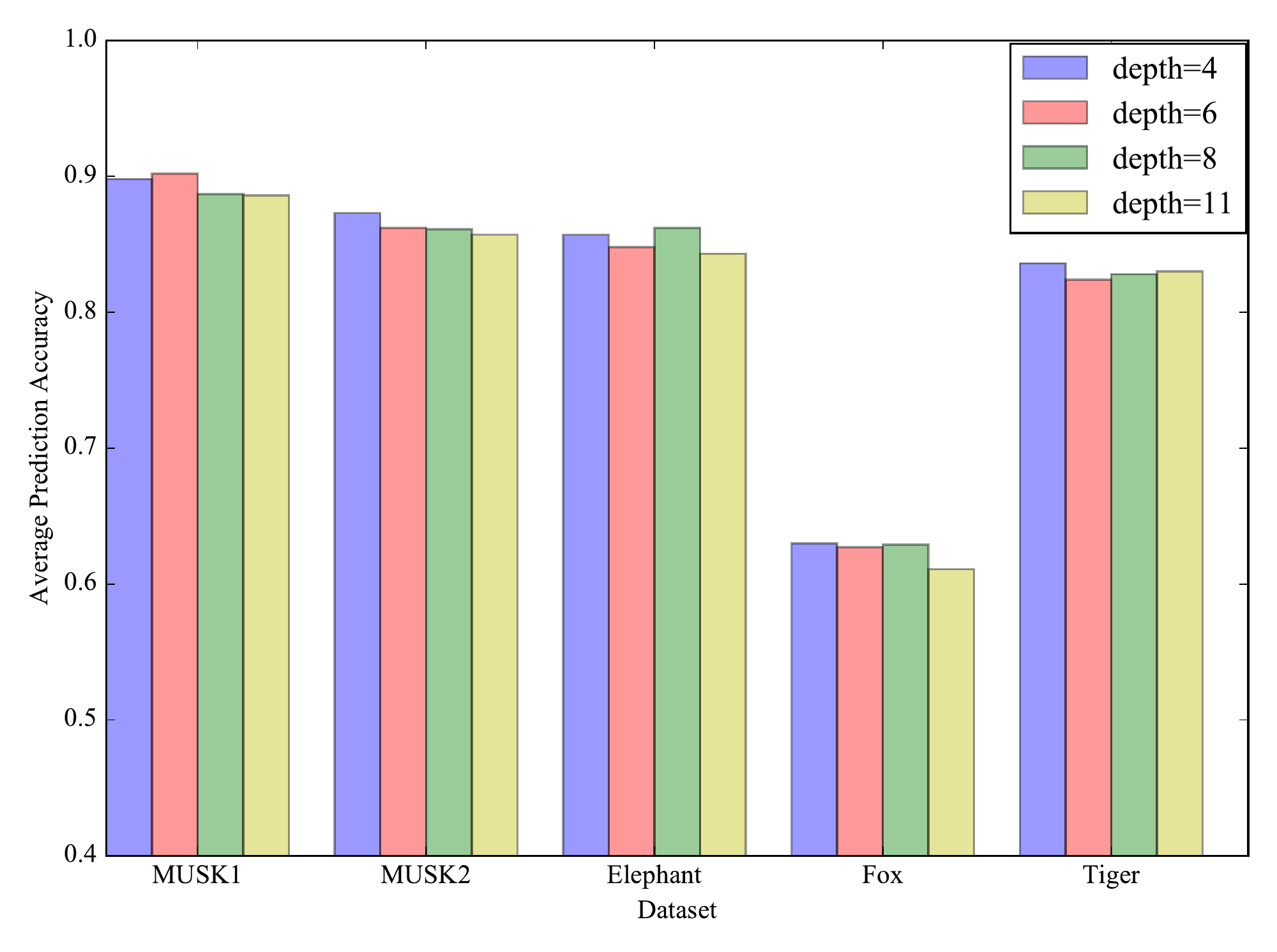}
  }
  \caption{Comparisons of depth for MI-Net with RC on five MIL benchmarks.}
  \label{fig:depth}
\end{figure}

\begin{figure}[t]
  \centerline{
    \includegraphics[width=8cm]{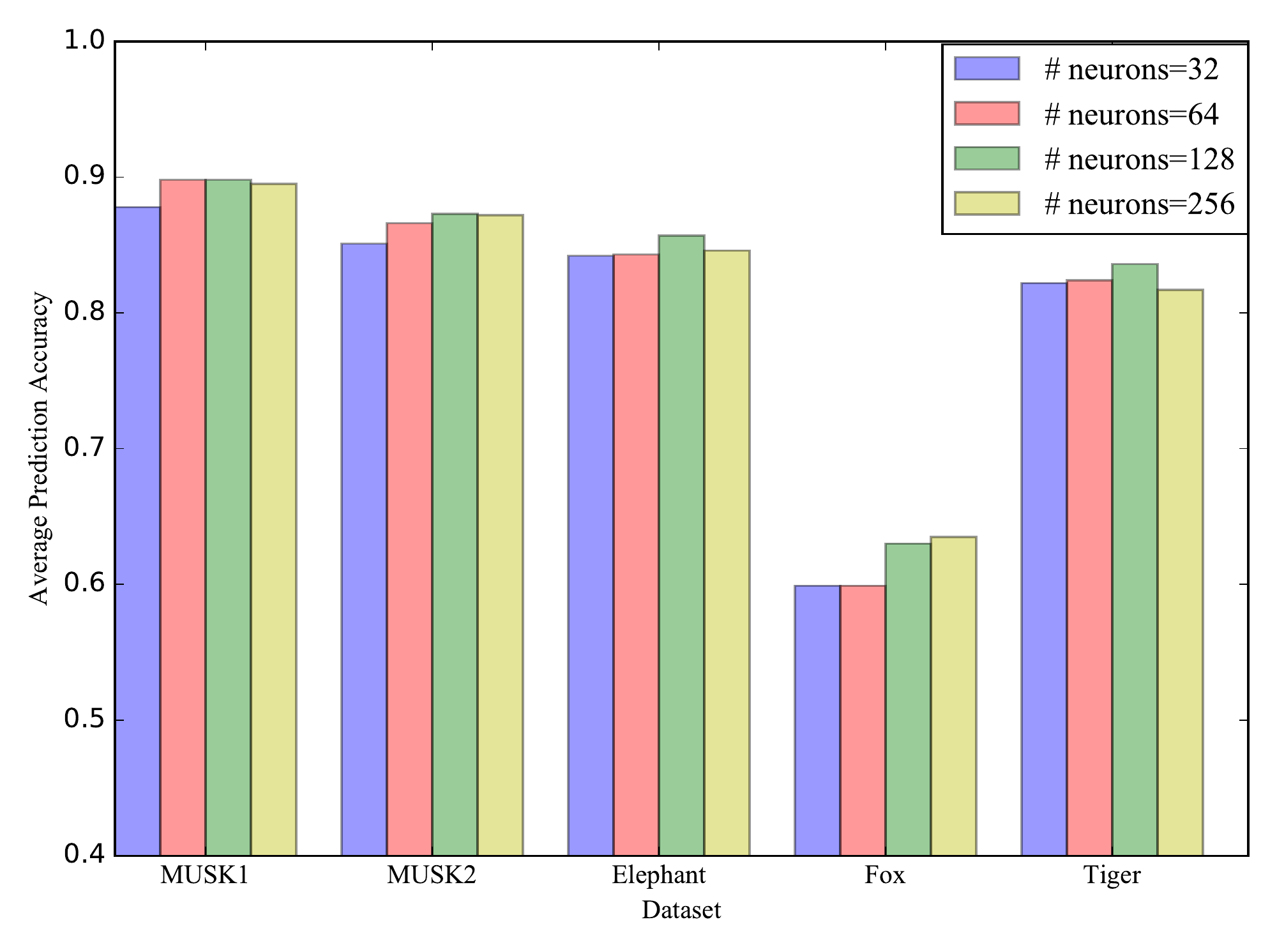}
  }
  \caption{Comparisons of width for MI-Net with RC on five MIL benchmarks.}
  \label{fig:width}
\end{figure}

\begin{table*}[ht]
  \caption{The influence of depth and width for MI-Net with DS on five MIL benchmarks, where numbers in brackets mean the number neurons for each fc layer.}
  \label{table:ds_structure}
  \centering
  \begin{tabular}{llllll}
    \toprule
     Structure	& MUSK1 & MUSK2 & Elephant & Fox & Tiger \\
    \midrule
	($256$, $256$, $256$, $1$)		& $\mathbf{0.898}$	& $0.853$	& $0.842$	& $0.629$	& $0.826$	\\
	($256$, $256$, $128$, $1$) 	& $0.881$	& $\mathbf{0.877}$	& $0.844$	& $0.602$	& $0.836$	\\
	($256$, $128$, $64$, $1$)		& $0.894$	& $0.874$	& $\mathbf{0.872}$	& $\mathbf{0.630}$	& $\mathbf{0.845}$	\\
	($128$, $128$, $128$, $1$)		& $0.887$	& $0.871$	& $0.840$	& $0.616$	& $0.836$	\\
	($128$, $128$, $64$, $1$)		& $0.866$	& $0.859$	& $0.845$	& $0.602$	& $0.836$	\\
	($64$, $64$, $64$, $1$)		& $0.891$	& $0.857$	& $0.861$	& $0.592$	& $0.824$  \\
	($256$, $256$, $128$, $128$, $64$, $1$)	& $0.892$	& $0.873$	& $0.844$	& $0.627$	& $0.835$	\\
	($256$, $256$, $256$, $256$, $256$, $1$)	& $0.884$	& $0.853$	& $0.838$	& $0.609$	& $0.835$	\\
    \bottomrule
  \end{tabular}
\end{table*}

As aforementioned,for mi-Net, MI-Net and MI-Net with its variations, the number of layers and neurons for each layer are fixed when training and testing. 
In table~\ref{table:exp1} and table~\ref{table:exp2}, the proposed network both have four fc layers and there are $256$, $128$, $64$, $1$ neurons for fc layers in MI-Net with DS respectively while $128$, $128$, $128$, $1$ in MI-Net with RC.
However, in deep learning, the deeper and wider neural network may get better performance. 
In this section, we will report the results of proposed MI-Net with DS and MI-Net with RC with different layer number and neuron number values on five MIL benchmarks, respectively.

The depth and width analysis results of MI-Net with DS on five MIL benchmarks are presented in Table~\ref{table:ds_structure}. 
Note that, the neuron number of last fc layer is fixed to 1 in order to output bag scores. 
As shown in Table~\ref{table:ds_structure}, MI-Net with DS can achieve the best performance in most cases when the depth is $4$, and each fc layer has $256$, $128$, $64$, $1$ neurons respectively. 
Although results of the deeper and wider network is superior to the shallower and thinner one on some datasets, the advantage of the deeper and wider network is not obvious to boost the performance.

As referred in Section~\ref{ssec:residual}, the neuron number of fc layers should be same value to build residual connections except for the last fc layer. 
Fixing the width of MI-Net with RC, we only change the depth of the network.
In Figure~\ref{fig:depth}, results on five MIL benchmarks are similar with the network deeper.
So the depth of MI-Net with RC is set to $4$ during discussing the influence of width on MI-Net with RC.
Figure~\ref{fig:width} illustrates that the wider network is not necessary to boost the performance.
In addition, MI-Net with RC may get worse performance when it is too thin.

This observation is not consistent with the performance of deeper and wider neural networks to solve other problems.
That may be related to limited training data and simple MIL pooling methods.

\section{Conclusion}
\label{sec:conclu}

In this work, we propose series of novel neural network frameworks for MIL.
Different from previous MIL networks, our method focuses on bag level representation learning instead of instance level label estimating.
Experiments show that our bag level networks show superior results on several MIL benchmarks compared with the instance level network.
Moreover, we intergrate the most popular deep learning tricks (deep supervision and residual connections) into our networks, which can boost the performance further.
What is more, our method only takes about $0.0003$ second for testing (forward) and $0.0008$ second for training (backward) per bag, which is very efficient.
According to these inspiring results, we believe that deep learning can also solve the traditional MIL problem well.
In the future, we would like to study how to develop more effective MIL pooling methods, and how to train deeper and wider networks for MIL with limited training data.

%In the paper, we develop neural networks to deal with data with more complex structure. Our finding is that deep supervision is essential in multiple instance networks. Besides, two fast and accurate MIL algorithms are proposed. In the future, we aim to more MIL pooling functions and develop more applications using multiple instance networks.

% or


%\appendices

%\section{}

% \section*{Acknowledgment}


\ifCLASSOPTIONcaptionsoff
  \newpage
\fi



%\bibliographystyle{IEEEtran}

\bibliographystyle{IEEEtran}
\bibliography{minet}

\begin{thebibliography}{10}
\providecommand{\url}[1]{#1}
\csname url@samestyle\endcsname
\providecommand{\newblock}{\relax}
\providecommand{\bibinfo}[2]{#2}
\providecommand{\BIBentrySTDinterwordspacing}{\spaceskip=0pt\relax}
\providecommand{\BIBentryALTinterwordstretchfactor}{4}
\providecommand{\BIBentryALTinterwordspacing}{\spaceskip=\fontdimen2\font plus
\BIBentryALTinterwordstretchfactor\fontdimen3\font minus
  \fontdimen4\font\relax}
\providecommand{\BIBforeignlanguage}[2]{{%
\expandafter\ifx\csname l@#1\endcsname\relax
\typeout{** WARNING: IEEEtran.bst: No hyphenation pattern has been}%
\typeout{** loaded for the language `#1'. Using the pattern for}%
\typeout{** the default language instead.}%
\else
\language=\csname l@#1\endcsname
\fi
#2}}
\providecommand{\BIBdecl}{\relax}
\BIBdecl

\bibitem{dietterich1997solving}
T.~G. Dietterich, R.~H. Lathrop, and T.~Lozano-P{\'e}rez, ``Solving the
  multiple instance problem with axis-parallel rectangles,'' \emph{Artificial
  Intelligence}, vol.~89, no.~1, pp. 31--71, 1997.

\bibitem{amores2013multiple}
J.~Amores, ``Multiple instance classification: Review, taxonomy and comparative
  study,'' \emph{Artificial Intelligence}, vol. 201, pp. 81--105, 2013.

\bibitem{hinton2006fast}
G.~Hinton, S.~Osindero, and Y.~W. Teh, ``A fast learning algorithm for deep
  belief nets,'' \emph{Neural computation}, vol.~18, no.~7, pp. 1527--1554,
  2006.

\bibitem{lecun1998gradient}
Y.~LeCun, L.~Bottou, Y.~Bengio, and P.~Haffner, ``Gradient-based learning
  applied to document recognition,'' \emph{Proceedings of the IEEE}, vol.~86,
  no.~11, pp. 2278--2324, 1998.

\bibitem{Ref:Krizhevsky2012}
A.~Krizhevsky, I.~Sutskever, and G.~E. Hinton, ``Imagenet classification with
  deep convolutional neural networks,'' in \emph{NIPS}, 2012, pp. 1097--1105.

\bibitem{williams1989learning}
R.~J. Williams and D.~Zipser, ``A learning algorithm for continually running
  fully recurrent neural networks,'' \emph{Neural computation}, vol.~1, no.~2,
  pp. 270--280, 1989.

\bibitem{hochreiter1997long}
S.~Hochreiter and J.~Schmidhuber, ``Long short-term memory,'' \emph{Neural
  computation}, vol.~9, no.~8, pp. 1735--1780, 1997.

\bibitem{ramon2000multi}
J.~Ramon and L.~De~Raedt, ``Multi instance neural networks,'' in
  \emph{Proceedings of the ICML-2000 workshop on attribute-value and relational
  learning}, 2000, pp. 53--60.

\bibitem{zhou2002neural}
Z.-H. Zhou and M.-L. Zhang, ``Neural networks for multi-instance learning,'' in
  \emph{Proceedings of the International Conference on Intelligent Information
  Technology, Beijing, China}, 2002, pp. 455--459.

\bibitem{andrews2002support}
S.~Andrews, I.~Tsochantaridis, and T.~Hofmann, ``Support vector machines for
  multiple-instance learning,'' in \emph{NIPS}, 2002, pp. 561--568.

\bibitem{srivastava2014dropout}
N.~Srivastava, G.~Hinton, A.~Krizhevsky, I.~Sutskever, and R.~Salakhutdinov,
  ``Dropout: A simple way to prevent neural networks from overfitting,''
  \emph{JMLR}, vol.~15, no.~1, pp. 1929--1958, 2014.

\bibitem{nair2010rectified}
V.~Nair and G.~Hinton, ``Rectified linear units improve restricted boltzmann
  machines,'' in \emph{ICML}, 2010, pp. 807--814.

\bibitem{lee2015deeply}
C.~Y. Lee, S.~Xie, P.~Gallagher, Z.~Zhang, and Z.~Tu, ``Deeply-{S}upervised
  {N}ets,'' in \emph{AISTATS}, 2015, pp. 562--570.

\bibitem{he2015deep}
K.~He, X.~Zhang, S.~Ren, and J.~Sun, ``Deep residual learning for image
  recognition,'' \emph{arXiv preprint arXiv:1512.03385}, 2015.

\bibitem{zhang2004improve}
M.-L. Zhang and Z.-H. Zhou, ``Improve multi-instance neural networks through
  feature selection,'' \emph{Neural Processing Letters}, vol.~19, no.~1, pp.
  1--10, 2004.

\bibitem{zhang2004ensembles}
M.~Zhang and Z.~Zhou, ``Ensembles of multi-instance neural networks,'' in
  \emph{International Conference on Intelligent Information Processing}.\hskip
  1em plus 0.5em minus 0.4em\relax Springer, 2004, pp. 471--474.

\bibitem{Sanchez2013}
J.~S{\'{a}}nchez, F.~Perronnin, T.~Mensink, and J.~J. Verbeek, ``Image
  classification with the {F}isher {V}ector: Theory and practice,''
  \emph{IJCV}, vol. 105, no.~3, pp. 222--245, 2013.

\bibitem{Ref:Wu2015}
J.~Wu, Y.~Yu, C.~Huang, and K.~Yu, ``Deep multiple instance learning for image
  classification and auto-annotation,'' in \emph{CVPR}, 2015, pp. 3460--3469.

\bibitem{pinheiro2015image}
P.~O. Pinheiro and R.~Collobert, ``From image-level to pixel-level labeling
  with convolutional networks,'' in \emph{CVPR}, 2015, pp. 1713--1721.

\bibitem{wang2015relaxed}
X.~Wang, Z.~Zhu, C.~Yao, and X.~Bai, ``Relaxed multiple-instance {SVM} with
  application to object discovery,'' in \emph{ICCV}, 2015, pp. 1224--1232.

\bibitem{Ref:Deng2009}
J.~Deng, W.~Dong, R.~Socher, L.~J. Li, K.~Li, and L.~Fei-Fei, ``Imagenet: A
  large-scale hierarchical image database,'' in \emph{CVPR}, 2009, pp.
  248--255.

\bibitem{glorot2011deep}
X.~Glorot, A.~Bordes, and Y.~Bengio, ``Deep sparse rectifier neural networks.''
  in \emph{Aistats}, vol.~15, no. 106, 2011, p. 275.

\bibitem{boyd2004convex}
S.~Boyd and L.~Vandenberghe, \emph{Convex optimization}.\hskip 1em plus 0.5em
  minus 0.4em\relax Cambridge university press, 2004.

\bibitem{zhang2001dd}
Q.~Zhang and S.~A. Goldman, ``{EM-DD}: An improved multiple-instance learning
  technique,'' in \emph{NIPS}, 2001, pp. 1073--1080.

\bibitem{gartner2002multi}
T.~G{\"a}rtner, P.~A. Flach, A.~Kowalczyk, and A.~J. Smola, ``Multi-instance
  kernels,'' in \emph{ICML}, vol.~2, 2002, pp. 179--186.

\bibitem{zhou2009multi}
Z.~H. Zhou, Y.~Y. Sun, and Y.~F. Li, ``Multi-instance learning by treating
  instances as non-iid samples,'' in \emph{ICML}, 2009, pp. 1249--1256.

\bibitem{wei2016scalable}
X.~S. Wei, J.~Wu, and Z.~H. Zhou, ``Scalable algorithms for multi-instance
  learning,'' \emph{IEEE Transactions on Neural Networks and Learning Systems},
  vol.~PP, no.~99, pp. 1--13, 2016.

\bibitem{glorot2010understanding}
X.~Glorot and Y.~Bengio, ``Understanding the difficulty of training deep
  feedforward neural networks,'' in \emph{AISTATS}, 2010, pp. 249--256.

\bibitem{chollet2015keras}
F.~Chollet, ``Keras,'' \url{https://github.com/fchollet/keras}, 2015.

\end{thebibliography}

\end{document}